\documentclass[runningheads]{llncs}
\usepackage{graphicx}
\usepackage[usenames,dvipsnames]{color}
\usepackage{amsmath}
\usepackage[table,xcdraw]{xcolor}
\usepackage{xurl}
\usepackage{lineno}

\usepackage[acronym,nomain,nonumberlist]{glossaries}
\makeglossaries

\newacronym{vc}{VC}{Visual Cortex}
\newacronym{ffa}{FFA}{Fase Fusiform Area}
\newacronym{lvc}{LVC}{Low Visual Cortex}
\newacronym{hvc}{HVC}{High Visual Cortex}
\newacronym{it}{IT}{Inferior Temporal Lobe}
\newacronym{gan}{GAN}{Generative Adversarial Network}
\newacronym{vae}{VAE}{Variational Autoencoder}
\newacronym{cnn}{CNN}{Convolutional Neural Network}

\begin{document}
\title{Semantic Brain Decoding: from fMRI to conceptually similar image reconstruction of visual stimuli}
\titlerunning{Semantic visual reconstruction from brain}
%
\author{Matteo Ferrante\inst{1}\and
Tommaso Boccato\inst{1} \and Luca Passamonti\inst{2}\and
Nicola Toschi\inst{1,3}}
\authorrunning{Ferrante et al.}
%
\institute{Department of Biomedicine and Prevention, University of Rome, Tor Vergata (IT) \and
Istituto di Bioimmagini e Fisiologia Molecolare, CNR, Milano \and Martinos Center for Biomedical Imaging, MGH and Harvard Medical School
\email{\{matteo.ferrante,tommaso.boccato, nicola.toschi\}@uniroma2.it}
\email{\{luca.passamonti\}@cnr.it} }

\maketitle              
\begin{abstract}
Brain decoding is a field of computational neuroscience that uses measurable brain activity to infer mental states or internal representations of perceptual inputs. We propose a novel approach to brain decoding that relies on semantic and contextual similarity. We employ an fMRI dataset of where vision of natural images was employed as stimuli and create a deep learning decoding pipeline inspired by the existence of both bottom-up and top-down processes in human vision. We train a linear brain-to-feature model to map fMRI activity features to visual stimuli features, assuming that the brain projects visual information onto a space that is homeomorphic to the latent space represented by the last convolutional layer of a pretrained convolutional neural network, which typically collects a variety of semantic features that summarize and highlight similarities and differences between concepts. These features are then categorized in the latent space using a nearest-neighbor strategy, and the results are used to condition a generative latent diffusion model to create novel images. From fMRI data only, we produce reconstructions of visual stimuli that match the original content very well, surpassing the state of the art in previous literature. We evaluate our work and obtain good results using a quantitative semantic metric (Wu-Palmer similarity metric over the WordNet lexicon, average = 0.57). We also perform a human evaluation experiment intended to reproduce the multiplicity of conscious and unconscious criteria that humans use to evaluate image similarity. This resulted in correct evaluation in over $80\%$ of the test set.

\end{abstract}
\section{Introduction}

\begin{figure}[h]
    \centering
    \includegraphics[width=\textwidth]{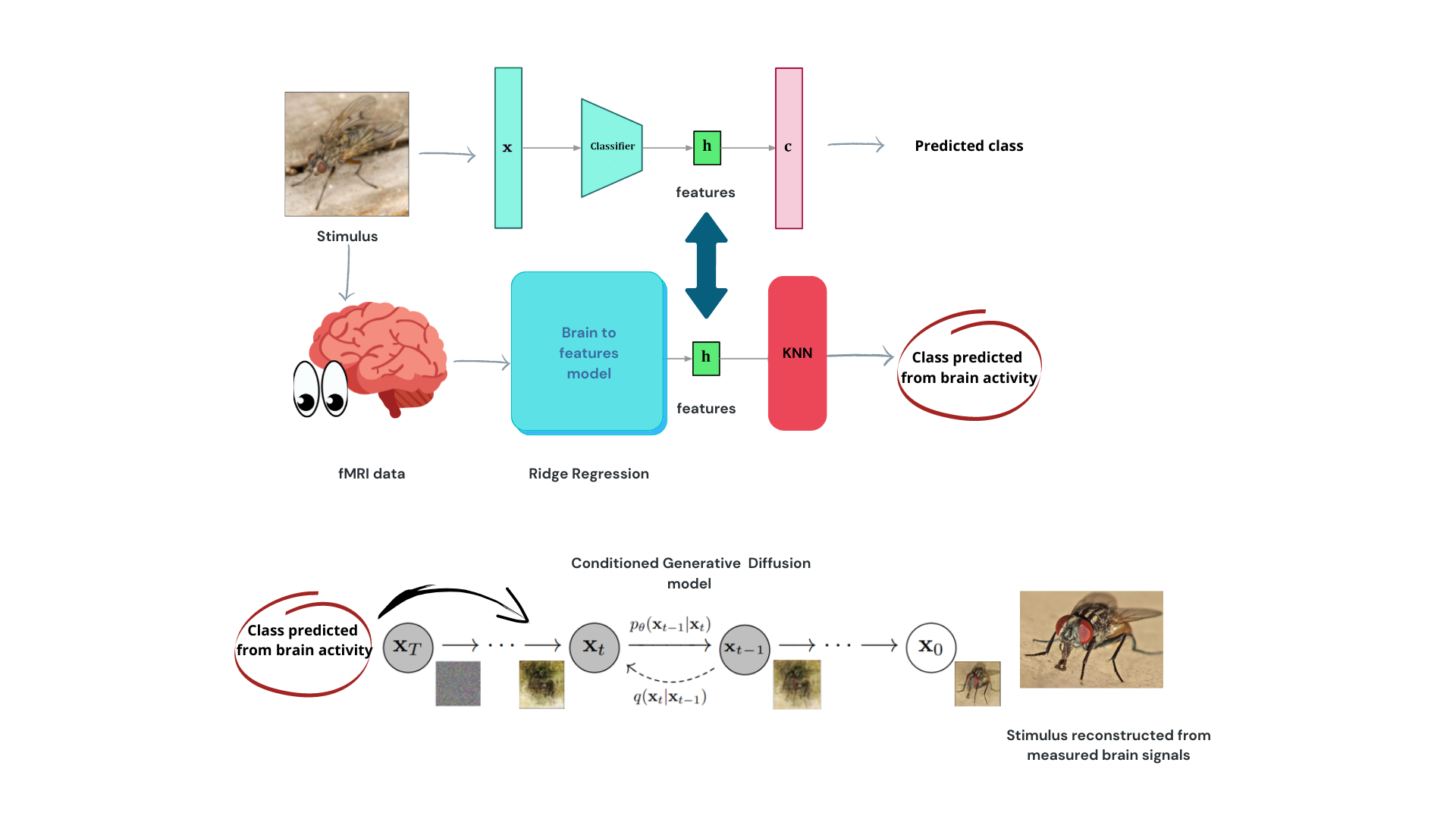}
    \caption{Our proposed architecture. According to our hypothesis, the brain processes information by extracting visual features from images and projecting them onto a latent semantic space similar to the one formed by a convolutional neural network (CNN), termed "classifier" in this figure (in this paper we employed the ResNet50 architecture) when trained for object categorization. We developed a regression model that maps fMRI brain data to the CNN’s latent space and used a k-nearest-neighbor (kNN) method to predict the related classes. Finally, we conditioned a latent diffusion model to generate novel images that are semantically similar to the visual stimuli from the predicted classes.}
    \label{fig:scheme}
\end{figure}

\subsection{Background and Motivation}

Brain decoding attempts to infer internal representations of perceptual stimuli from measurable brain activity. Isolated attempts have been made to use deep learning to 1) identify complex brain data patterns and 2) reconstruct the stimuli that have generated such patterns using noninvasive neuromonitoring data such as functional magnetic resonance imaging (fMRI) or electroencephalography (EEG) \cite{zafar_decoding_2015}. While these activities are in very early stages, they also carry great promise for the development of novel strategies to diagnose and treat neurological or neuropsychiatric conditions. However, such endeavors carry many challenges. Noninvasive data, for example, have lower temporal or spatial resolution than that of neural firing, resulting in a potential upper limit on the granularity of information that may be retrieved. The latter is also degraded by physiological noise and signal/image artifacts.

\subsection{Visual Cortex and Pathways}

Vision has been extensively studied along with its brain representations (i.e., the visual cortex). The latter are organized hierarchically into sections that respond to specific stimuli (commonly termed V1, V2, V3, V4, and the lower and upper visual cortices). Simple visual inputs tend to elicit V1 responses, while V2 responds to texture, color, and more complex outlines. There is strong evidence that information flows from the visual cortex (VC) to the rest of the brain through two separate routes, the \textit{what} and \textit{where} pathways \cite{bar_visual_2004,ungerleider_what_1994,gross_visual_1972,goodale_separate_1992}. The \textit{what} pathway connects the VC to the inferior temporal lobe (IT) and is involved in object recognition, whereas the \textit{where} pathway connects the VC to the parietal lobe and is primarily involved in movement and position recognition.

\subsection{Semantic Representations in the Brain}

In vision, the bottom-up information extraction described above is accompanied by a top-down mechanism \cite{gilbert_brain_2007} where semantic prior knowledge of the world is exploited to create internal representations of external stimuli. This results in a combination of context-given prediction and purely external signals relayed from the retina to the brain. According to the ‘hub-and-spoke’ theory of semantic representation, conceptual knowledge arises from the progressive learning of the statistical regularities of our multi-sensorial experiences. In other words, we learn how to recognize an ever-changing environment by systematically linking apparently separate aspects of our experiences (e.g., color, motion, sounds, sensory-motor actions associated with an object, etc.) that tend to co-occur. Such learning processes transform a sensory ‘cacophony’ into a coherent, context-specific, and behaviorally-relevant semantic representation of the stimuli.

\subsection{Semantic Cognition and Modality-Specific Brain Areas}

The brain mechanisms underlying semantic cognition have not been fully elucidated, but the prevailing hypothesis suggests that modality-specific brain areas, also known as the ‘spokes’ (e.g., visual cortices, auditory cortices, motor areas, emotional systems), interact via a central and a-modal ‘hub’ region (the anterior temporal lobe) to form conceptual knowledge. This process shapes the semantic representation through various experiences, such as visual, auditory, verbal, and tactile, and critically promotes the ability to generalize across different items and variable contexts. Interestingly, there are indications of the existence of a continuous semantic space representation \cite{huth_continuous_2012} in the human brain.
Though the structure and topology of this putative semantic space have been poorly investigated, there is evidence that fMRI data from occipital brain regions collected during a visual task can be linked to features learned by a convolutional neural network (CNN) \cite{lindsay_convolutional_2021}, with a particular focus on the early and middle CNN layers.

\subsection{Decoding Visual Stimuli from fMRI Data}

In this paper, we tackle the problem of decoding (i.e., reconstructing) visual stimuli (images) from fMRI data only, by proposing the hypothesis that deep convolutional layers can operate as a proxy for parts of the brain that extract semantic features from images. We propose a cascade of deep learning models that builds convincing semantic reconstructions of the stimulus presented at acquisition time. It is important to note that the aim of this paper is not to create exact reconstructions of the images presented under fMRI. Instead, our objectives are to either a) generate realistic visual representations that capture the main concepts contained in the original stimulus, or b) create synthetic images that can trigger similar brain activity when employed as stimuli. Achieving either of these results can pave the way for a more general understanding of cognitive-visual information storage and retrieval.

\section{Related Work}

\subsection{Reconstructing Information from fMRI Data}

In recent years, several attempts have been made to reconstruct information from noninvasively acquired brain data, particularly fMRI data. This has been fueled by the increasing availability of public datasets, advances in computational power, and more sophisticated nonlinear analytic approaches, such as deep neural networks. While challenges related to signal-to-noise ratio (SNR), duration of acquisition session, and HRF variability remain, fMRI appears capable of extracting useful information in a wide range of situations and tasks, including vision and visual stimulus classification.

\subsection{Existing Approaches and Challenges}

Various modeling frameworks have been employed in brain decoding literature, where the input is usually preprocessed fMRI time series. These data are referred to as “fMRI data”, “fMRI patterns”, and “fMRI activations”, terms used interchangeably in this paper. Existing approaches to brain decoding include:

\begin{itemize}
\item Variational autoencoder with a generative adversarial component (VAE-GAN) for encoding latent representations of human faces \cite{faces}.
\item Sparse linear regression over preprocessed fMRI data for predicting features extracted by multiple early convolutional layers from a pretrained CNN \cite{horikawa_generic_2017}.
\item An adversarial strategy employing a generator and discriminator to differentiate between real and reconstructed images, further improved by a perceptual loss and a comparator network \cite{shen_end--end_nodate}.
\item A dual VAEGAN consisting of two linked variational autoencoders for representing both stimuli and fMRI patterns \cite{ren_reconstructing_2019}.
\item An unsupervised technique using two encoders and two decoders learning separately how to reconstruct fMRI data and stimuli, bound by a supervised loss \cite{gaziv_self-supervised_2022}.
\item Optimizing pretrained architectures' latent spaces, such as BigBiGAN \cite{mozafari_reconstructing_2020} and IC-GAN \cite{ozcelik_reconstruction_2022}, to reconstruct high-quality images from fMRI patterns.

\item \cite{chen2022seeing} performed a direct estimation of the latent space of a latent diffusion model from fMRI data, employing a pre-trained autoencoder to reduce the dimensionality of fMRI representations. By combining the HCP \cite{HCP} (1,200 subjects) and GOD datasets, they achieved a substantial sample size to learn self-supervised representations and fine-tune them for inferring the latent representations of images with limited labeled pairs. Our work is closely related due to the utilization of the same GOD dataset and latent diffusion models for image reconstruction; however, our main distinction lies in the development of an ad-hoc pipeline to address the small sample size of the GOD dataset independently, whereas they relied on external fMRI acquisitions to learn self-supervised representations.
\item Latent diffusion models have been recently employed as image generators in \cite{Takagi2022.11.18.517004} and \cite{ozcelik2023braindiffuser}, where the authors utilized the Natural Scenes Dataset \cite{NSDDataset}, containing 70,000 images acquired with a 7T scanner. This extensive dataset significantly enhances the quality and quantity of input data for brain decoding tasks. In the first study, the authors directly optimized the latent space of the diffusion model, while in the second, an initial guess image reconstruction was obtained by mapping fMRI data into the latent space of a deep variational autoencoder trained for image reconstruction. These initial guesses encompass information about shape, color, and pose of images, and can be combined with predicted conditioning in latent diffusion models through image-to-image pipelines to improve reconstruction quality. An important distinction between our work and these studies is that our dataset's sample size is nearly two orders of magnitude smaller, and was acquired on a 3T rather than a 7T acquisition. This implies that the performances of our model have been obtained with a small fraction of the information and of the signal-to-noise ratio available to other models.

\end{itemize}

\subsection{Focus on Semantic Content}

Most of the research in brain decoding has focused on extracting either low-level visual stimulus characteristics or reconstructing whole images in pixel space. While these studies capture forms, colors, or images that look similar to the original stimuli, reconstructions are often blurred and mix elements from unrelated concepts. In this paper, we focus on context, i.e., the semantic content of presented stimuli, with the aim of reconstructing images that resemble the original ones and can elicit the same fMRI activity. We hypothesize that this approach may add ecological relevance to our findings in terms of future applications for understanding visual information representation in the brain.
\section{Methods}
In this section, we describe the implementation aspects of our study. We used Python 3.9 along with the PyTorch and scikit-learn libraries to develop our models. The experiments were conducted on a server equipped with two Intel Xeon Gold processors, 512 GB RAM, and an NVIDIA A6000 GPU with 48 GB RAM. Our code is available at \url{https://github.com/matteoferrante/semantic-brain-decoding}, and the preprocessed data can be accessed at \url{https://figshare.com/articles/dataset/Generic_Object_Decoding/7387130}. Unprocessed fMRI data is available at \url{https://openneuro.org/datasets/ds001246/versions/1.2.1}.

\subsection{Data and Preprocessing}
\label{sec:data}

We utilize the publicly accessible Generic Object Decoding (GOD) dataset \cite{horikawa_generic_2017}, which comprises fMRI data from 5 subjects who participated in either an image presentation experiment or an imagery experiment. The GOD dataset has been instrumental in developing previous brain decoding models and is emerging as a valuable benchmark for decoding visual stimuli from fMRI data. All visual stimuli in the GOD dataset originate from the ImageNet database (\url{http://www.image-net.org/}, Fall 2011 release), which is categorized into various classes, including animals (e.g., "goldfish," "swarm," and "tiger") and objects (e.g., "airplane," "hat," or "knife").

The image presentation experiment involved separate training and test sessions. In the training session, 1,200 images from 150 object categories (8 images per category) were presented once. In the test session, 50 images from 50 object categories (1 image per category) were shown 35 times each. Each stimulus was displayed for nine seconds. No overlap existed between the categories of training and test images. In this dataset, a single fMRI acquisition is called a "run," with 24 runs for training images and 35 runs for testing images performed for each subject. The fMRI protocol was based on an EPI sequence with $TR=3000$ ms, $TE=30$ ms, flip angle=$80°$, and a voxel size of 3 $mm^3$.

Data were preprocessed in native subject space by performing 3D motion correction, linear trend removal, and coregistration to a high-resolution common anatomical template. Reference masks for the visual cortex (VC) and several other brain areas, such as the face fusiform area (FFA), the high VC (HVC), and the low VC (LVC), were provided for each subject. In this study, we used data extracted from the VC (approximately 4,500 voxels per subject) as our input space. The data were normalized runwise, ensuring each voxel-specific timeseries had a zero mean and unit variance. Subsequently, data were averaged over time using nonoverlapping 9-second windows and effectively shifted forward by 3 seconds (i.e., three volumes per average, corresponding to the length of a stimulus presentation). This process helped reduce complexity and account for delays induced by the hemodynamic response function (HRF) convolution.

\subsection{Subject-specific Brain Activity Models}
We developed individual models for each subject to decode their brain activity, as intersubject functional variability could be greater than the impact we aim to extract. Our hypothesis is that the brain processes sensory input in the VC to extract relevant features from images for object recognition, employing a hierarchical approach similar to convolutional neural networks (CNNs).

We propose a linear mapping between processed fMRI data and the last convolutional layer of the ResNet50 \cite{resnet} architecture, trained on the ImageNet dataset. The objective is to find the optimal weights $W$ that minimize the regularized loss described in Eq. \eqref{eq:ridgeloss}:

\begin{equation}
\label{eq:ridgeloss}
\min(|Wx(s)-f(s)|^2 + \lambda|W|^2)
\end{equation}

Here, $s$ represents the image/stimulus presented during the experiment, $f$ is the neural network that projects $s$ into the latent space, and $x(s)$ is the preprocessed brain activity associated with viewing the stimulus. $W$ maps fMRI data into image features in the latent space generated by ResNet50. $\lambda$ is a hyperparameter for $L2$ regularization on the weights. We optimized $\lambda$ using a $90-10$ training/validation split and grid search.

Subsequently, we generated the conditioning for the generative model that synthesizes the final output. We used ResNet50 to compute the latent representation of a subset of 500K randomly selected ImageNet pictures and stored their latent representation and ground truth labels. From the image features $\tilde{h}=Wx(s)$ predicted from brain activity, we identified the five nearest neighbors in the latent space and used their labels as candidates for classification.

This strategy accounts for the poor signal-to-noise ratio in fMRI data and the limited dataset size. Assuming that similar semantic concepts lead to similar features within the ResNet50 latent space, the features generated by our brain-to-features model (ridge regression) are likely to be close to concepts semantically close to the target.

\subsection{Bottom-up and Top-down Processes}
\label{sec:bottom_up_top_down}

We now discuss the combination of predicted features that simulate the bottom-up process in vision (where the brain computes stimuli) and the use of a nearest-neighbor-based algorithm to mimic top-down connections that modulate the signal we perceive according to our knowledge of the world. We also address the domain adaptation technique employed to predict the test set features from brain activity, the use of latent diffusion models as image generators, and the evaluation of semantic content through metrics such as the Wu-Palmer distance. In this study, we combine predicted features to simulate the bottom-up process in vision, where the brain computes stimuli, while using a nearest-neighbor-based algorithm to mimic top-down connections that modulate the signal we perceive according to our knowledge of the world \cite{DDPM,GaNvsDM}.

\subsection{Domain Adaptation Technique}
\label{sec:domain_adaptation}

There is no overlap between training and test categories in the GOD dataset, and test images are displayed numerous times to achieve a higher SNR. Since the brain-to-feature model is trained using training data, we employ a simple domain adaptation technique to predict test set features from brain activity, which involves replacing the mean and standard deviation of predicted features from the test set with those from the training set \cite{DDPM,GaNvsDM}.

\subsection{Latent Diffusion Models as Image Generators}
\label{sec:latent_diffusion_models}

To generate images (i.e., reconstruct visual stimuli), we rely on a powerful, recent pretrained image generator belonging to the family of denoising probabilistic diffusion models \cite{DDPM}. Diffusion models are generative architectures that learn how to reverse a diffusion process, which in this context refers to the progressive addition of Gaussian noise to an image. This family of models is far more robust in training than other generative models, such as generative adversarial networks (GANs), and has greater mode coverage \cite{GaNvsDM}.

\subsection{Evaluating Semantic Content}
\label{sec:evaluating_semantic_content}

Our primary objective is to produce images that are close (in a semantic space) to the real visual stimuli shown to participants during the fMRI experiment. We create two metrics specifically designed to evaluate the quality of the generated images. First, we use the Wu-Palmer distance metric \cite{wupalmer} between the real and predicted classes in the WordNet lexicon to estimate a quantifiable measure of semantic similarity. This is a well-established metric that measures the similarity of two different nodes (i.e., synsets) in the WordNet graph and can be computed as described in Eq \eqref{eq:wupalmer}, where $s$ is the similarity metric, $lcs$ stands for “least common subsumer” and is a function that returns the deepest common ancestor in the taxonomy between the two synsets $s1,s2$ and $depth$ is a function that computes the depth in the graph. This metric is bounded in the interval $[0,1]$, where higher values mean that two synsets are more similar. A simplified graphical representation of the WordNet subgraph is shown in Fig. \ref{fig:wupalmer} along with some examples of Wu-Palmer distances.

Additionally, to quantify the performance of our model we used the Fréchet inception distance score (FID) \cite{fid}. In this metric, two sets of images (real and generated) are compared as multivariate Gaussian distributions in the feature space of a pretrained neural network (InceptionV3). While this metric mainly measures the quality of images both in terms of feature distribution and of visual quality, and thus is likely to mainly reflect the performance of our image generation model, the projection into the feature space of InceptionV3 is likely to also endow this metric with elements of semantic similarity. We used the torchmetrics \cite{torchmetrics} library implementation and compared the images generated for each subject with those used as stimuli. It should be noted that the main evaluation criterion used in brain decoding literature is a visual comparison between reconstructions of the same image across models.

\begin{equation}
    \label{eq:wupalmer}
    s_{wup}= \frac{depth(lcs(s1,s2))}{depth(s1)+depth(s2)}
\end{equation}

In addition to the Wu-Palmer distance metric, we conducted a human evaluation to assess the semantic similarity of the reconstructed images.

\subsection{Human Evaluation of Image Reconstructions}

\begin{figure}

    \centering
    \includegraphics[width=\textwidth]{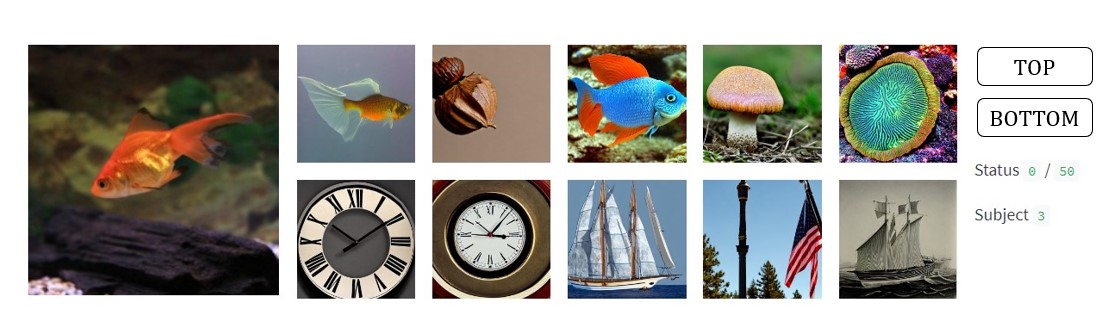}
    \caption{Example taken from the local human assessment local web page. The target image is presented on the left. The subject is instructed to assess the overall resemblance of the original stimulus (left) to the 5 images in the top and bottom rows on the right and to pick “TOP” or “BOTTOM ” accordingly.}
    \label{fig:humanevaluation}
\end{figure}

We designed a human evaluation paradigm as follows. A local web page was created, which displayed the original image alongside five model-generated reconstructions in one row and five random reconstructions in another row (Fig. \ref{fig:humanevaluation}). Volunteers were instructed to examine the similarities between the images and select the row (first or second) that appeared closest to the original image. To minimize priming, the row positions were continuously randomized between "top" and "bottom."

Seven observers (5 males, 2 females, aged 25-33, with normal eyesight) participated in this evaluation, covering all subjects in the GOD dataset. Each observer assessed the 50 images in the test set and a common random subset of 50 images from the training set, resulting in a total of 350 evaluations. When performing this task, the human observers likely focused on various elements, including broad features like shapes and colors, as well as more semantically related aspects, such as "wild animals" or "furnishings." We believe this natural flexibility in judgment is relevant to our study, as the model utilizes features extracted by a classifier trained on the ImageNet dataset. These features can represent different levels of complexity based on the difficulty of the task, and similar comparison operations might be performed by our brains in everyday life. To further minimize priming, the row positions were continuously randomized between "top" and "bottom."


\section{Results}

\subsection{Visual Comparison and Qualitative Results}

\begin{figure}[t]
    \centering
    \includegraphics[width=\textwidth]{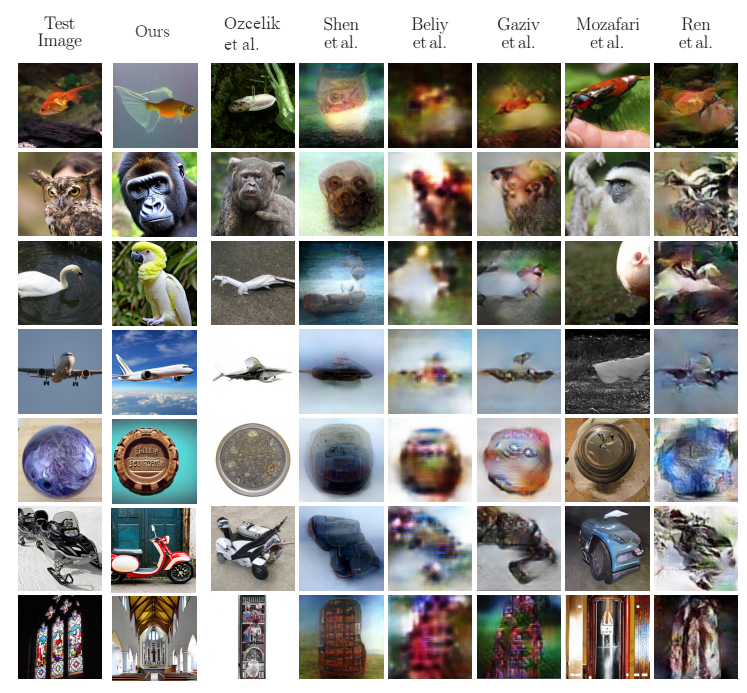}
    \caption{Comparison with previous approaches in brain decoding of visual stimuli over the GOD dataset. The first column shows original images used as stimuli, while other columns are reconstructions from different works. Our results are depicted in the second column.}
    \label{fig:comparison}
\end{figure}

The primary objective of our study is to generate images that are realistic reconstructions of visual inputs that semantically match the target image, which is the image used as a stimulus in the fMRI experiment. Fig. \ref{fig:comparison} presents a comparison with state-of-the-art reconstruction approaches over the same dataset, demonstrating qualitative differences between our approach and others. Our diffusion model generates images that are crisp and sharp and convey clear and specific content, which helps recognize similarities between images and distinguish between failed and successful semantic reconstructions.

We propose a paradigm shift in our approach to reconstruction. Rather than focusing on obtaining accurate reconstructions in pixel space, we aim to produce novel images that are semantically and contextually as close to the target visual stimulus as possible. For instance, reconstructions of "fish" and "airplane" (see Fig. \ref{fig:comparison}, first and fourth rows, with the first column showing original images and the second column showing our reconstructions) are among our best results, as they clearly portray the same concepts as the original image. Other images that match the stimulus on a semantic level, such as the swan that is reconstructed as a parrot (both birds), the snowmobile that is reconstructed as a motorbike (both vehicles), or the colorful church window reconstructed as a church, are instances of visuals that match the content and context without being exact pixelwise reconstructions.

\begin{figure}[!h]
    \centering
    \includegraphics[width=0.48\textwidth]{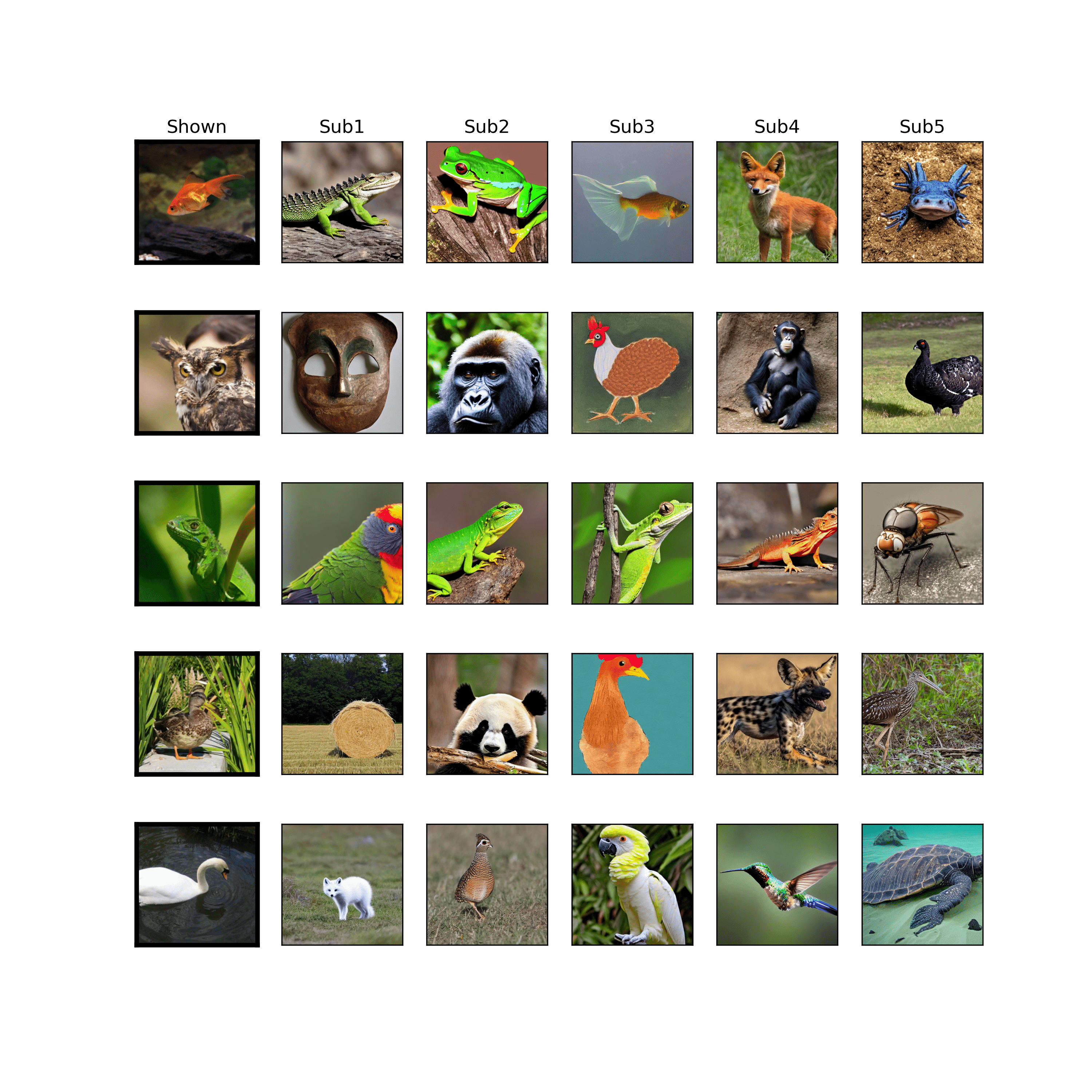}
    \includegraphics[width=0.48\textwidth]{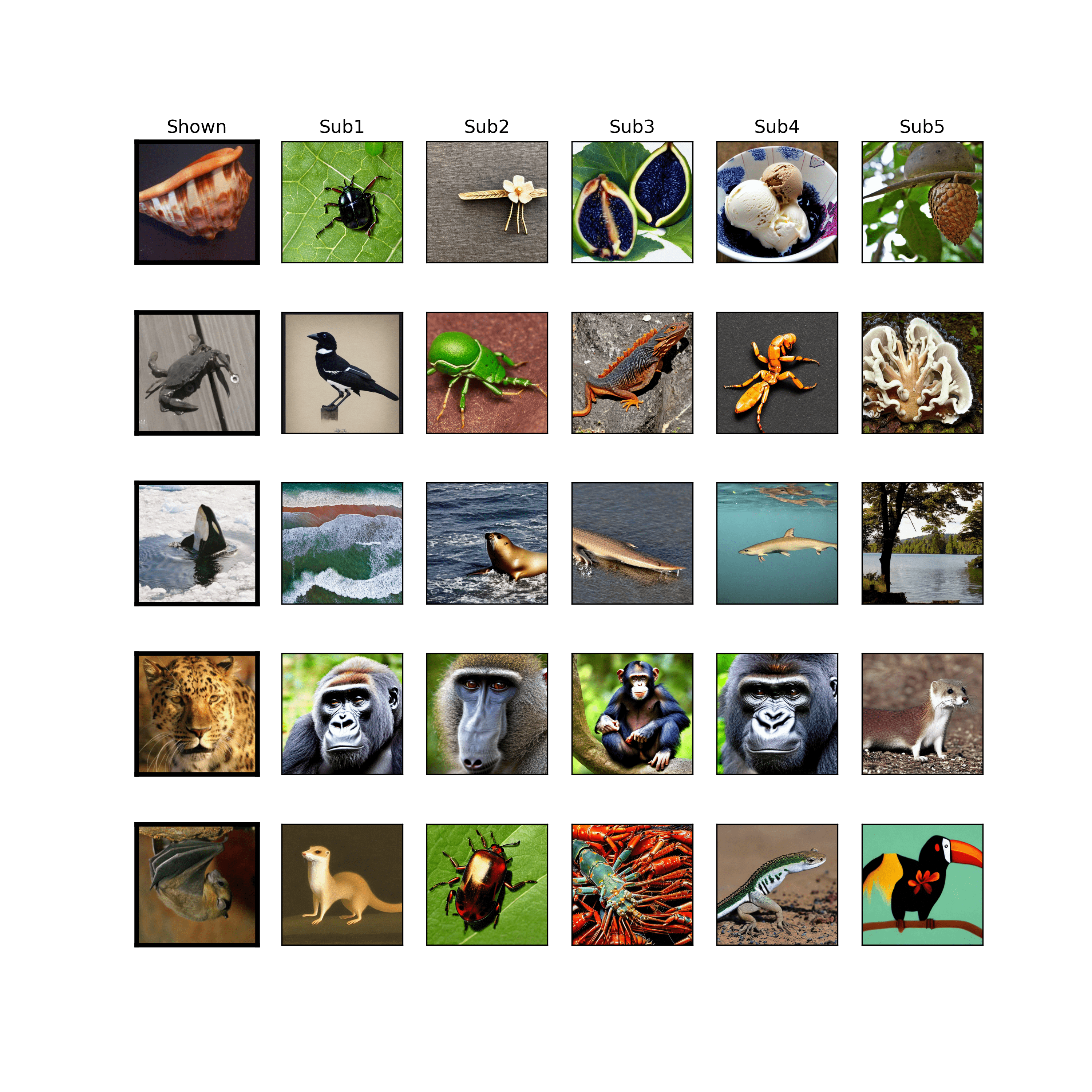}
    
    \caption{Some examples Examples of our semantic reconstructions over the test set. Left columns: original image stimulus shown to the subjects under fMRI. Other columns: semantic reconstructions for each subject in the GOD dataset.}
    \label{fig:examples}
\end{figure}

Fig. \ref{fig:examples} and the Supplementary Material show additional reconstruction examples for all subjects. One can see that our model provides a plausible reconstruction that matches the original at some contextual level in the majority of cases, although with a natural degree of variation that reflects the breadth of possible semantic similarities.

\subsection{Quantitative semantic distance}

We achieved a FID score of $10.58 \pm 1.95$ (mean $\pm$ standard deviation, test set) and an average Wu-Palmer distance of $0.811\pm0.204$ over the training set and $0.571\pm0.157$ over the test set (Fig \ref{fig:wupalmer-panel}). It is important to note that the images in the test set correspond to categories that do not overlap with those in the training set. Therefore, the quality of prediction in the test set is determined by the number of features shared by the two sets. However, there is a notable factor of similarity between original and generated images, even in the test dataset, suggesting that the brain-to-feature model can estimate semantic features related to groups of objects, such as wings, fur, and buildings, correctly. This result holds even though the model is trained on data with different categories and data distribution. In other words, our model performs well in spite of the non-overlap between training and test categories. While a simple classifier would likely not be able to generalize to this particular test set, our model performs well and demonstrates the potential for brain decoding to generalize to new categories and data distributions.


\subsection{Human Evaluation}

Humans perform well in complex assessments with wide criteria and can naturally examine images at numerous levels of semantic information as well as shapes, colors, and many more. Fig. \ref{fig:humanevaluation} and Table \ref{tab:human} show the results of human evaluation for both the training and test sets. On average, human observers selected the images generated from the model (as opposed to the randomly generated images) in $95\pm 3\%$ of the cases for images from the training set and in $81\pm 4\%$ of the cases for images from the test set. In all cases, human observers chose the model-generated images far more frequently than what would have been the chance level, supporting the hypothesis that our computational approach can correctly capture various semantic features of the images in a manner that corresponds well to the way the human brain evaluates this type of content and context.

\begin{figure}[!h]
    \centering
    \includegraphics[width=0.9\textwidth]{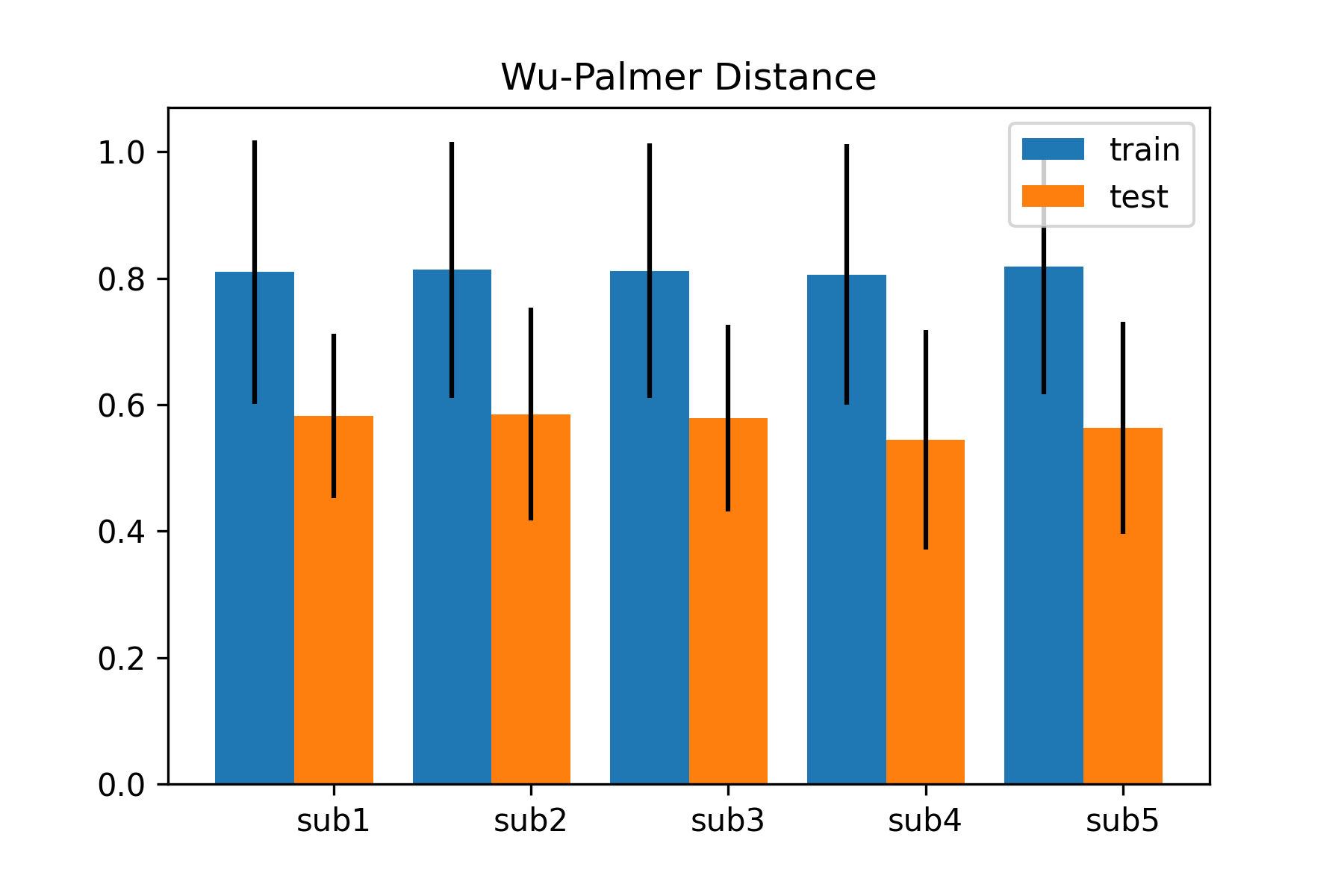}
    \caption{Wu-Palmer distances (mean +/- s.d.)  between original image stimuli shown to the subjects under fMRI for all subjects for both training (blue) and test (orange) sets.}
    \label{fig:res-wupalmer}
\end{figure}

\begin{figure}[!h]
    \centering
    \includegraphics[width=\textwidth]{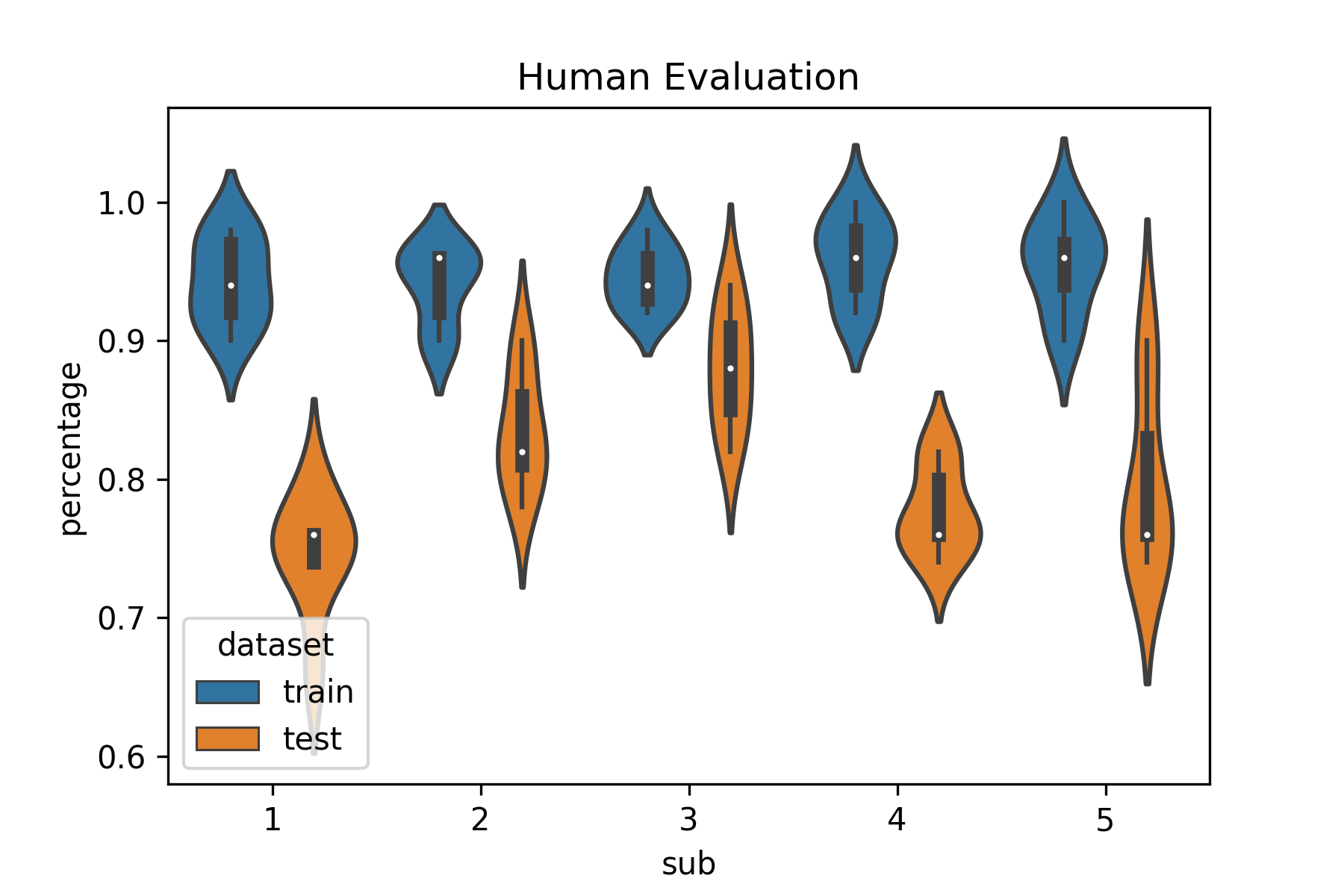}
    \caption{Human evaluation: Rate of selection (mean $\pm$ std) of images  generated by our model versus random images from human evaluators for images in the training (blue) and testing (orange) set s.}
    \label{fig:human}
\end{figure}

\begin{table}[]

\centering
\resizebox{0.8\textwidth}{!}{%
\begin{tabular}{l|ll}
\textbf{Subject} & \multicolumn{1}{l|}{\textbf{\begin{tabular}[c]{@{}l@{}}Human Evaluation\\ Training Dataset\end{tabular}}} & \textbf{\begin{tabular}[c]{@{}l@{}}Human Evaluation \\ Test Dataset\end{tabular}} \\ \hline
\textit{1}       & 0.960 $\pm$ 0.031                                                                                         & 0.778 $\pm$ 0.031                                                                 \\
\textit{2}       & 0.945 $\pm$ 0.022                                                                                         & 0.880 $\pm$ 0.043                                                                  \\
\textit{3}       & 0.940 $\pm$ 0.028                                                                                         & 0.834 $\pm$ 0.043                                                                 \\
\textit{4}       & 0.943 $\pm$ 0.031                                                                                         & 0.745 $\pm$ 0.042                                                                 \\
\textit{5}       & 0.954 $\pm$ 0.031                                                                                         & 0.797 $\pm$ 0.059        
                                        
\end{tabular}
}
\caption{Results of human evaluation. Rate of selection of images generated by our model model versus random images from human evaluators.}
\label{tab:human}

\end{table}

\section{Discussion}

\subsection{Developing the Brain-to-Feature Model and Reconstruction Pipeline}
Grounded in the assumption that fMRI data from the VC during a visual task can be used as a proxy for the last layer of a convolutional neural network (CNN) trained for image classification, we developed a brain-to-feature model. This model is a trained ridge regression between fMRI and image features extracted from the original visual stimuli images through ResNet50, establishing univocal relationships between fMRI data and the ResNet50 features \cite{lindsay_convolutional_2021,matsuo_generating_2016,huth_continuous_2012}.

We subsequently employed a nearest neighbor-like technique to map these features into object "categories." These categories were then used to condition a pretrained latent diffusion model to produce novel images from text prompts corresponding to the synset name of the related WordNet class. Our reconstruction pipeline incorporates these hypotheses through the mapping between fMRI and ResNet50 latent space, the use of the k-nearest neighbors algorithm, and reliance on a powerful image generator.

\subsection{Bottom-Up and Top-Down Processes in Vision}
Our brain-to-feature model represents the bottom-up process in vision, a rapid initial estimate of relevant features. This estimate is refined by our top-down approach, represented by the choice of the nearest neighbor in the latent space to condition the generative model. This component of our architecture is supported by prior knowledge of the world, contained in the ResNet50 latent space representation of a subset of the ImageNet database. This, in turn, allows us to evaluate the "distance" between concepts.

\subsection{Evaluating Performance Through Semantic-Related Measures}
We assessed our work both qualitatively (visually) and quantitatively through semantic-related measures. We employed the Wu-Palmer distance to analyze similarities between concepts in the WordNet lexicon, discovering a good average similarity. Additionally, we included an assessment of the contextual distance between original and reconstructed stimuli by naïve human observers to allow for additional flexibility and human-like semantic evaluation. Our results suggested that the model performed well in selecting relevant features and producing images closer to the original than any other image.

\subsection{Reconstruction Performance and Categories}
We found that with all assessment techniques, reconstructed images are rarely noticeably distant from the target, similar to the results reported in \cite{huth_continuous_2012}. Specifically, original images of animals generated reconstructions that accurately depicted other animals, with striking accuracy in high-level features such as "species". Similarly, original images of non-animated objects, such as vehicles, exhibited comparable behavior, giving rise to accurate renderings of planes, motorbikes, tractors, and carriages. While a similar behavior occurred for most visual stimuli, some categories appeared to be “misunderstood” by our model, such as the cowboy hat or the guitar (see Supplementary Material). In this context, it is possible that the traits associated with certain test images are underrepresented in the training set, increasing the difficulty of capturing relevant semantics.

\subsection{Brain and Deep Learning Models}
Our brain can be thought of as a prediction machine that utilizes past knowledge in the form of top-down processing of external inputs. We found that in the VC, this might produce a feature space that is homeomorphic to the latent space of a CNN. Notably, a linear (ridge regression) model was sufficient to achieve convincing reconstruction results. These findings are in line with evidence that deep learning models and brain activity prompted by language converge \cite{caucheteux_brains_2022,caucheteux_deep_2022,goldstein_shared_2022,matsuo_generating_2016} in terms of behavioral, physiological, and fMRI data, supporting our key hypothesis that context and semantics play a significant role in how we process sensory information. These ideas bear similarities to the concepts of attention-based deep learning models with convolutional layers.

\subsection{Semantic Cognition and Reconstruction}
Semantic cognition refers to a group of neuropsychological processes that sustain not only conceptual representation and formation but also the manipulation of semantic knowledge to influence context-relevant behavior. These brain mechanisms are thought to depend on a constant flow of top-down and bottom-up interactions between posterior and anterior areas, including occipito-temporal cortices and prefrontal networks. In the visual domain, the 'bottom-up' and 'top-down' interplay between multiple occipitotemporal cortices might allow the 'distillation' of a latent space of features that are believed to be at the ‘core’ of semantic representation. Our reconstruction approach, which utilized a combination of brain-to-feature and generative models, allowed us to recreate the original visual stimuli and obtain reconstructions of the images that surpass the state-of-the-art in the literature, particularly at the semantic level of reconstruction. This supports our approach's validity and its ability to mimic the way the human brain extracts, categorizes, and internally represents visually acquired information.
We employed a deep latent diffusion model to generate novel images that could evoke similar brain activity, featuring images with congruent semantic content. This capacity to synthesize images with precise content directly from brain activity lays the foundation for more advanced analyses and reconstructions. For instance, utilizing an image-to-image diffusion model that starts with an initial guess containing low-level aspects such as colors and shapes can lead to more accurate and plausible reconstructions.

\subsection{Neurobiological considerations}

It is important to note that our deep learning architecture is conceptually inspired by the current understanding of the neural mechanisms underlying semantic cognition. However, our model only employed fMRI data from a group of visual cortices (V1, V2, V3, V4) due to practical and computational considerations \cite{neural-basis-computation}. This choice does not deny the critical role of other brain regions, such as the anterior temporal lobe (ATL), in semantic cognition. The spoke-hub theory of semantic cognition clearly states that semantic cognition arises from the interplay of modality-specific (sensory, motor cortices) and a-modal regions (ATL, prefrontal cortex, etc.) \cite{spoke-hub-theory}. Future investigations could explore the role of other brain regions, such as the ATL, and determine whether the features extracted from those regions are superior to those of other regions of the brain when decoding the "mental states" associated with visual processing.

\subsection{Limitations}

The fMRI experiments used to collect the data were restricted in length because individuals need to be exposed to images slowly enough for the brain response to stabilize. As a result, the applicability of end-to-end deep learning algorithms is limited. In addition, because the categories in the training and test sets in the dataset we used do not overlap, the model's performance depends on the relationship between the fMRI data and image features in the training set. The assumption is that this relationship is sufficient to detect variations in unseen categories. Our model demonstrated good generalization capabilities, suggesting that semantic feature content, rather than precise train/test class overlap, may be predominant in determining performance. However, if the categories are highly dissimilar between the test and training set, it is conceivable that their essential properties are underrepresented in the training set, limiting the model's performance capabilities in the test . Also, it is important to note that we outperformed all models trained on the same 3T dataset, hence potentially widening the applicability of our methods to an extremely large number of centers wich do not have access to ultra-high-filed (7T or more) scanners.

Furthermore, there are numerous potential sources of error that can arise between the vision process and the generation of the image feature space. These include fMRI acquisition noise, bias in the feature space of the ResNet50 architecture, bias introduced by the limited sample size in the brain-to-feature model, and errors introduced by the conditioning algorithm. These circumstances can be responsible for cases where the performance of our model in reconstructing context is poor.

Additionally, mental attention may warp the semantic space in the human brain \cite{cukur_attention_2013}. When subjects become tired or bored during fMRI sessions, the encoded stimuli may change, introducing another source of variability that is not under experimental control. These limitations and sources of error should be taken into account when interpreting the results and considering future research directions.

\subsection{Future directions}

Future research could delve into the role of mental attention in semantic cognition and examine whether attentional states can modulate the distributed neural representations of semantic concepts in the visual cortex \cite{cukur_attention_2013}. Such investigations would contribute to our understanding of how attention influences decoding accuracy and the neural mechanisms underlying semantic cognition.
The growing availability of extensive open fMRI datasets will likely enable us to enhance brain decoding results using diffusion models as image generators, by conditioning these models in various ways. Interestingly, the majority of work in this field, including our own, currently focuses on subject-wise reconstruction. It would be intriguing to develop models capable of decoding intra-subject activity. This could pave the way for large-scale decoding on new subjects by merely fine-tuning a more extensive model, thus bypassing the need for lengthy fMRI acquisitions for each individual.
Another crucial next step is decoding imagery activity, reconstructing examples of images seen exclusively by the mind's eye.
Naturally, this raises ethical concerns regarding privacy and confidentiality, as decoding brain activity entails accessing an individual's internal mental state, potentially revealing sensitive information about their thoughts, emotions, and behavior. There is a risk that such information could be misused or disclosed without the person's consent, leading to privacy breaches. Ethical questions also arise concerning the accuracy of decoded images, which may produce a distorted version of a person's perception due to model imperfections.
Nonetheless, this type of research can lead to numerous beneficial applications. For instance, a completely new form of art could emerge from the interaction between the physics of fMRI acquisition, the artist's thoughts and perceptions, and the artificial intelligence used for decoding. This technology could also enable individuals with locked-in syndrome to communicate through images.

Moreover, future investigations could employ other brain imaging modalities, such as EEG or MEG, to investigate the temporal dynamics of the neural representations of semantic concepts and how they evolve over time during visual processing. Additionally, future studies could employ multi-modal data fusion methods to combine fMRI data with other modalities, such as behavioral data or natural language descriptions of visual stimuli, to gain a more comprehensive understanding of the neural basis of semantic cognition.

\section{Conclusions}

Our study proposes a pipeline to synthesize images that are conceptually and semantically similar to the original stimuli, starting from fMRI data only. We assume that measurable neural correlates can be linearly mapped onto the latent space of a convolutional neural network that represents a semantic description of the image. The overall objective is to replicate the way humans process information by combining bottom-up visual inputs with top-down cognitive descriptions of the environment, which is known to aid in "classification" processes in the brain.

We evaluated our reconstructions qualitatively and quantitatively and discovered a good Wu-Palmer similarity metric on the WordNet lexicon ($0.57\pm 0.15$) between true and predicted concepts, as well as high performance in the test set ($0.81\pm0.04$) when human observers evaluated the quality of our reconstructions in a double-blind process. Our work represents an improvement in the decoding of visual stimuli with respect to previous studies by including a semantic-based hypothesis in our reconstruction pipeline.

In summary, our study provides evidence that measurable neural correlates can be linearly mapped onto the latent space of a convolutional neural network to synthesize images that are conceptually and semantically similar to the original stimuli. The findings have implications for both cognitive neuroscience and artificial intelligence, as they shed light on the neural mechanisms underlying visual perception and suggest promising avenues for future research.

\section*{Acknowledgments}
Part of this work is supported by the EXPERIENCE project (European Union’s Horizon 2020 research and innovation program under grant agreement No. 101017727)
Matteo Ferrante is a Ph.D. student enrolled in the National PhD in Artificial Intelligence, XXXVII cycle, course on Health and life sciences, organized by Università Campus Bio-Medico di Roma.

%
%
%
\glsaddall

\printglossary[type=\acronymtype,title=Acronyms]

\bibliographystyle{splncs04}
\bibliography{brain_decoding}

\newpage
\section*{Supplementary Material}
Here are presented all reconstructions obtained by our model for the entire test set, for each subject. The first column shows the images used as stimuli, so the target of the reconstruction. The other columns show semantic reconstructions of the target image for all subjects.

\begin{figure}[h!]

\includegraphics[width=1\linewidth]{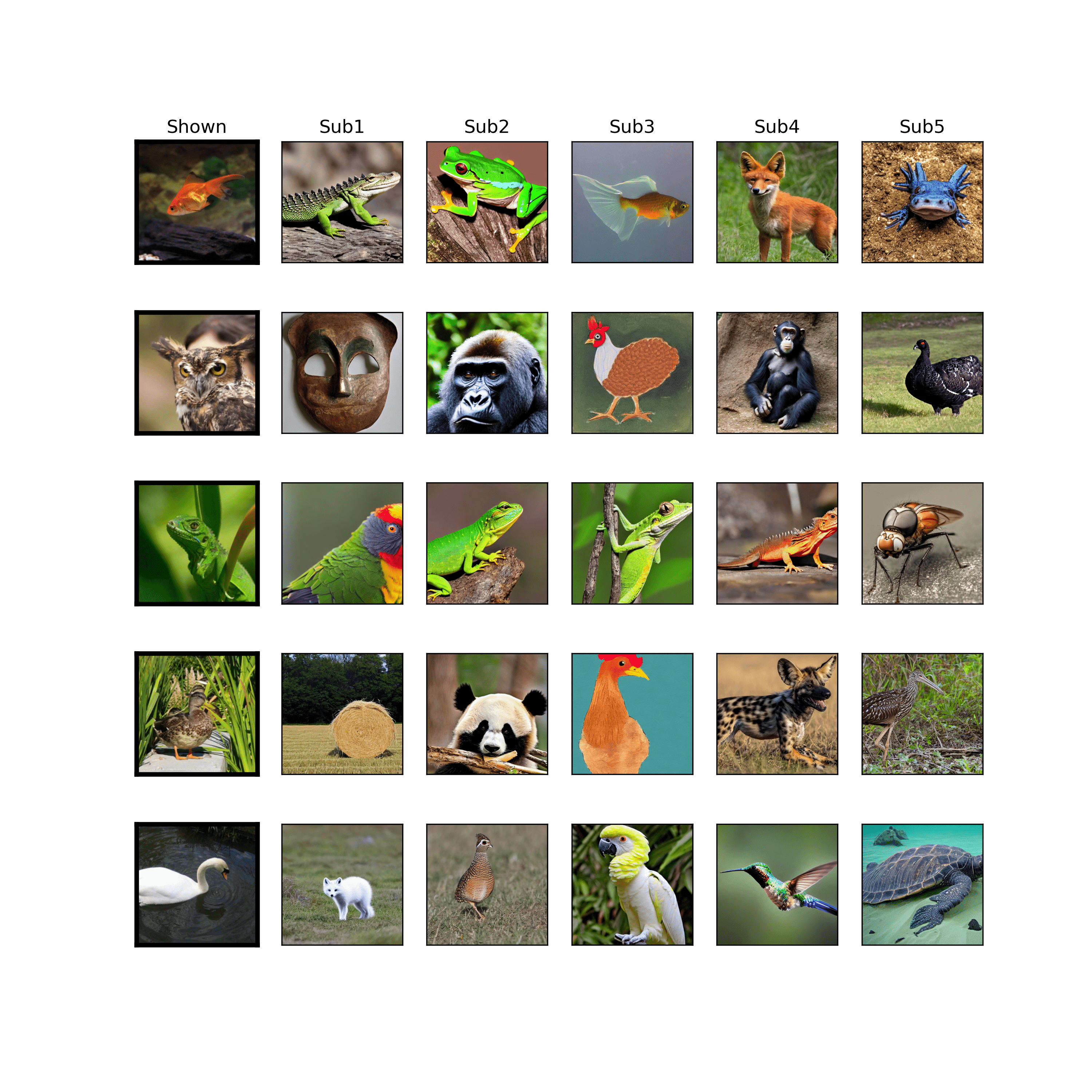}

 \caption{Examples of our semantic reconstructions over the test set. Left columns: original image stimulus shown to the subjects under fMRI. Other columns: semantic reconstructions for each subject in the GOD dataset.}
    \label{fig:examples-app-10}
\end{figure}

\begin{figure}[h!]

\includegraphics[width=1\linewidth]{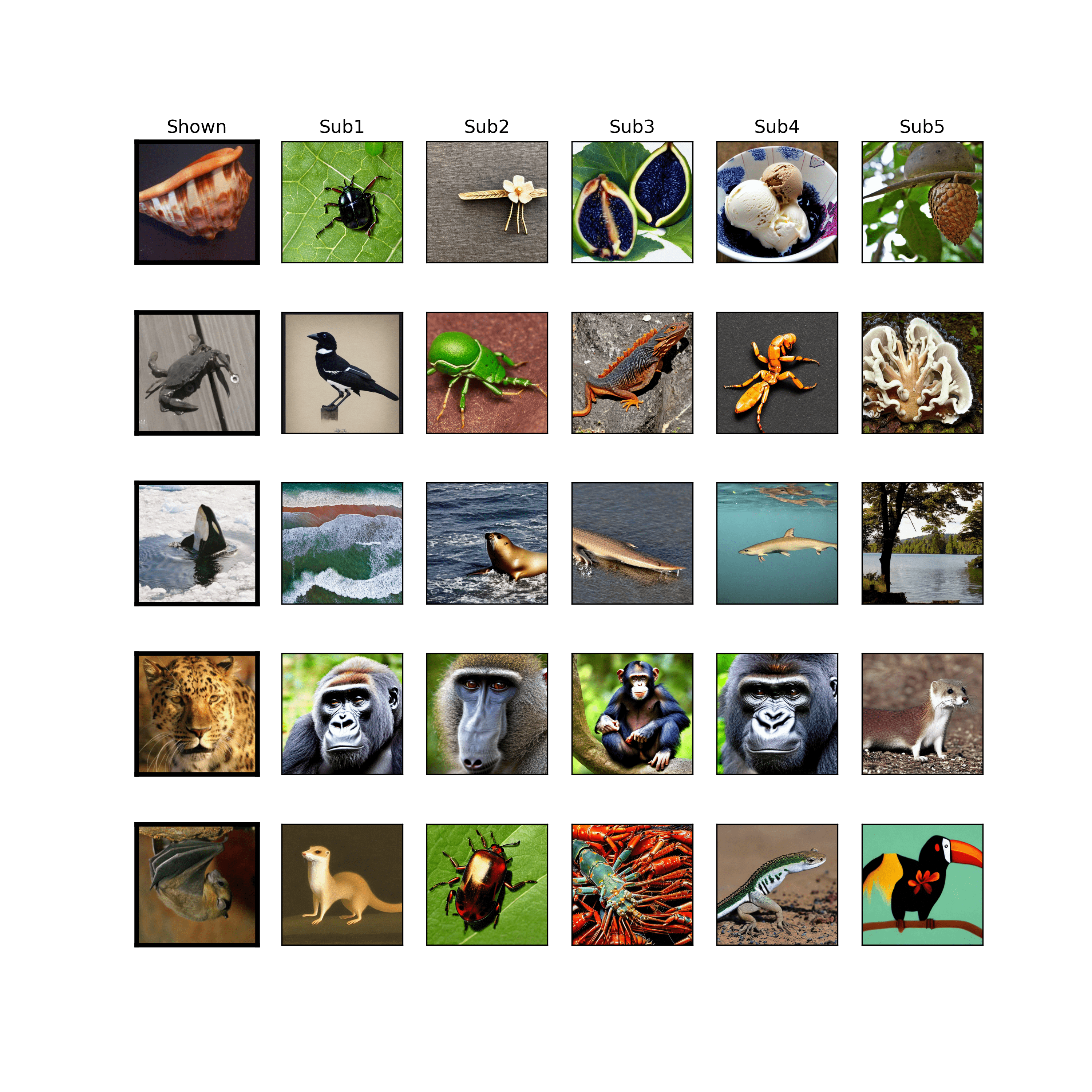}

 \caption{Examples of our semantic reconstructions over the test set. Left columns: original image stimulus shown to the subjects under fMRI. Other columns: semantic reconstructions for each subject in the GOD dataset.}
    \label{fig:examples-app-10}
\end{figure}

\begin{figure}[h!]
\includegraphics[width=1\linewidth]{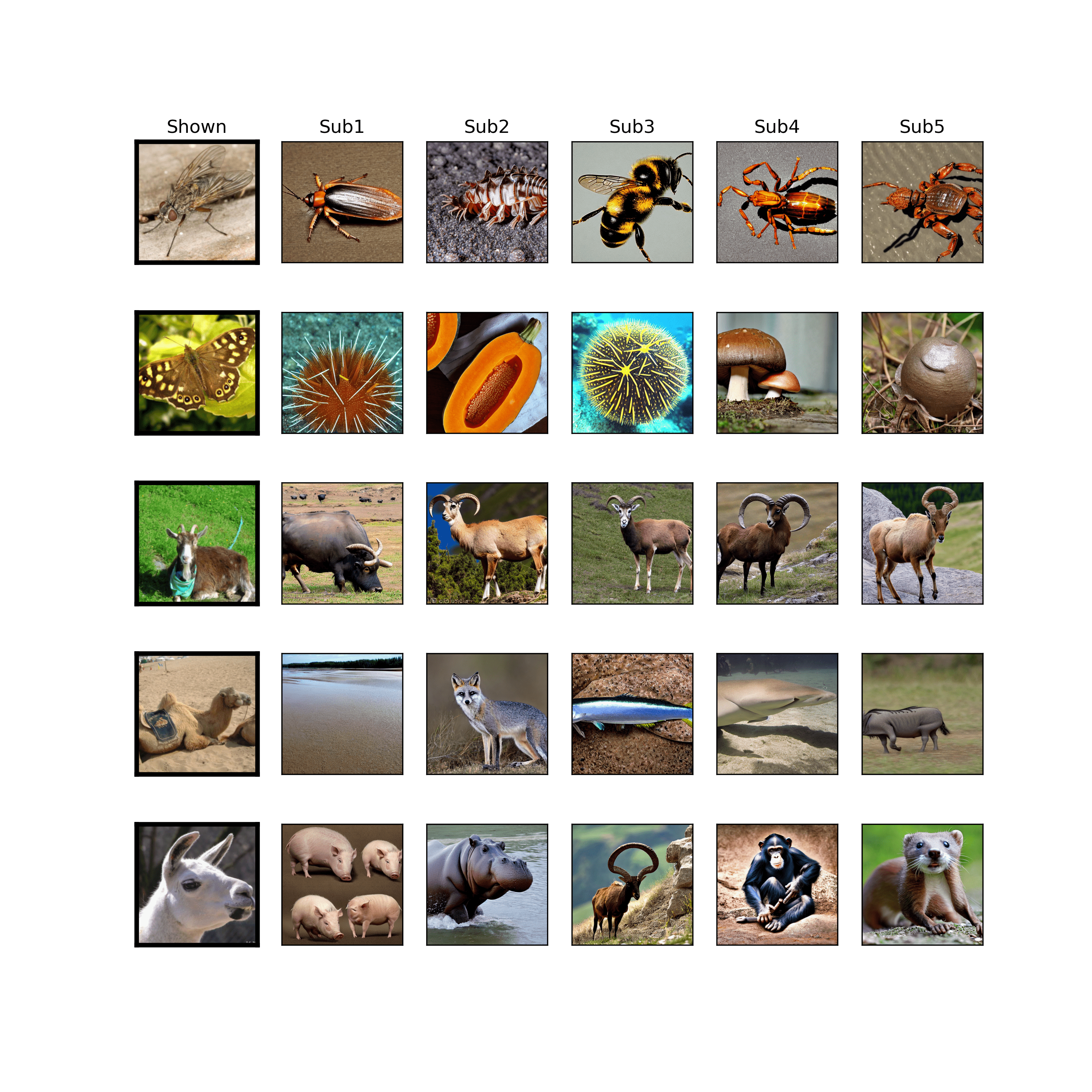}

 \caption{Examples of our semantic reconstructions over the test set. Left columns: original image stimulus shown to the subjects under fMRI. Other columns: semantic reconstructions for each subject in the GOD dataset.}
    \label{fig:examples-app-10}
\end{figure}

\begin{figure}[h!]
\includegraphics[width=1\linewidth]{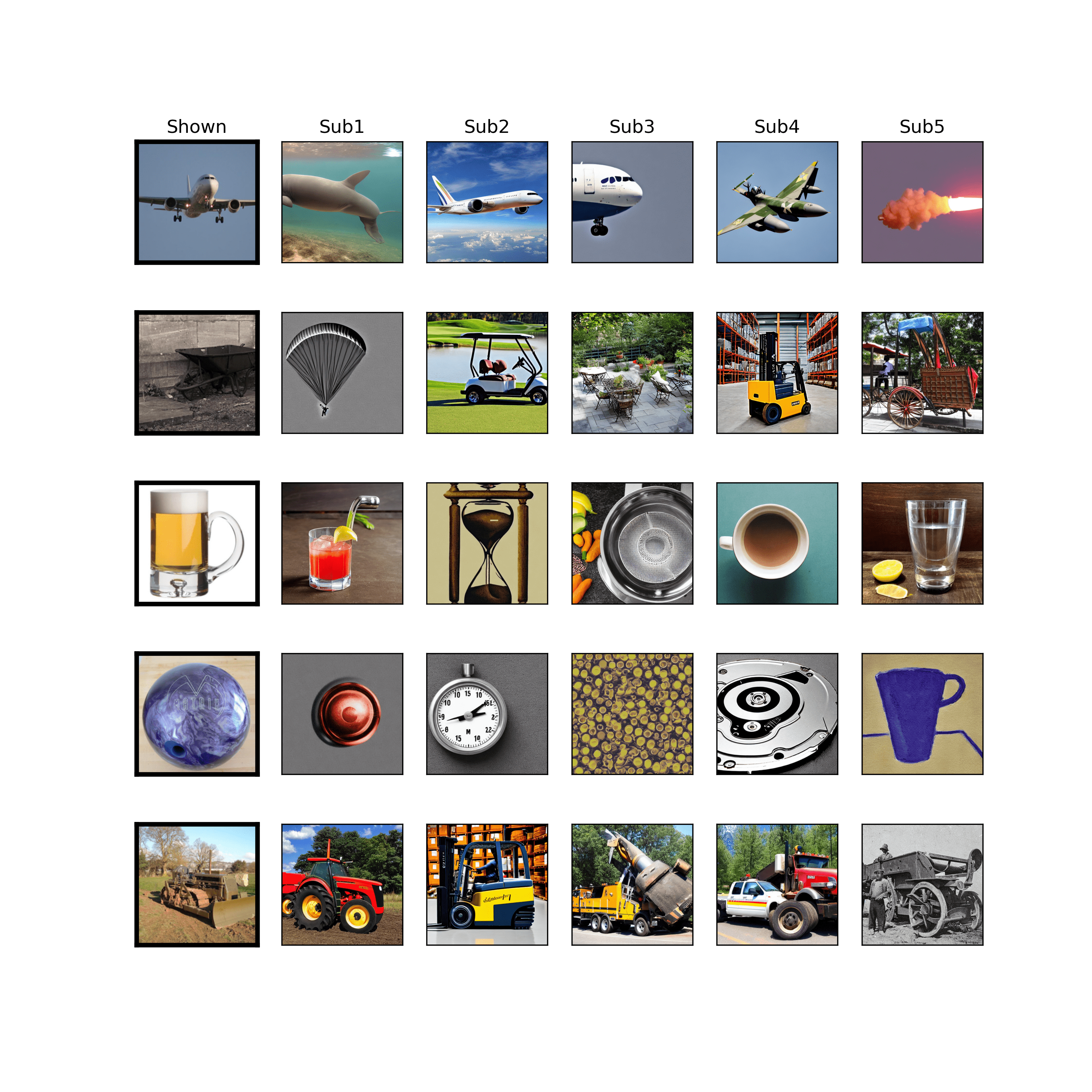}

 \caption{Examples of our semantic reconstructions over the test set. Left columns: original image stimulus shown to the subjects under fMRI. Other columns: semantic reconstructions for each subject in the GOD dataset.}
    \label{fig:examples-app-10}
\end{figure}

\begin{figure}[h!]
\includegraphics[width=1\linewidth]{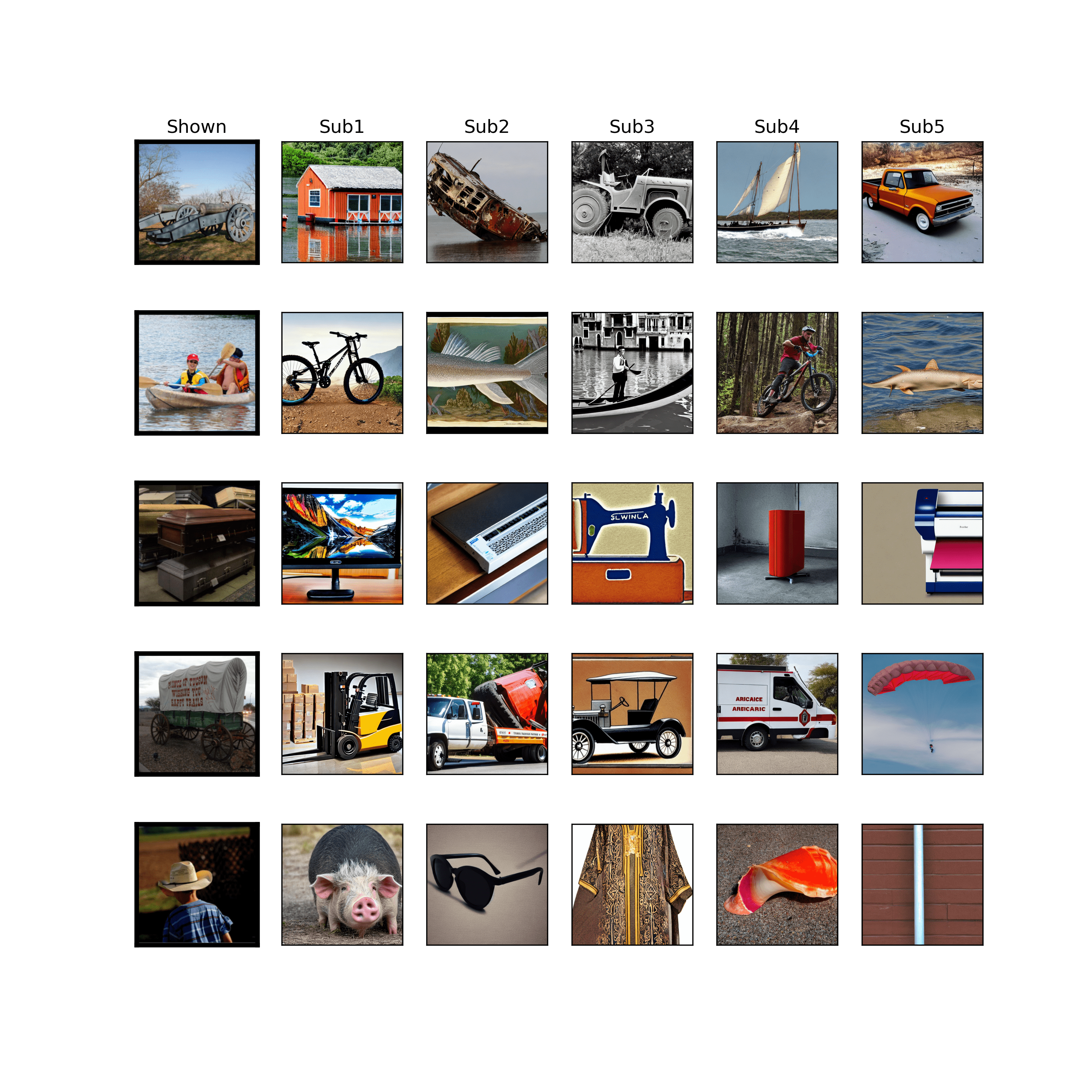}

 \caption{Examples of our semantic reconstructions over the test set. Left columns: original image stimulus shown to the subjects under fMRI. Other columns: semantic reconstructions for each subject in the GOD dataset.}
    \label{fig:examples-app-10}
\end{figure}

\begin{figure}[h!]
\includegraphics[width=1\linewidth]{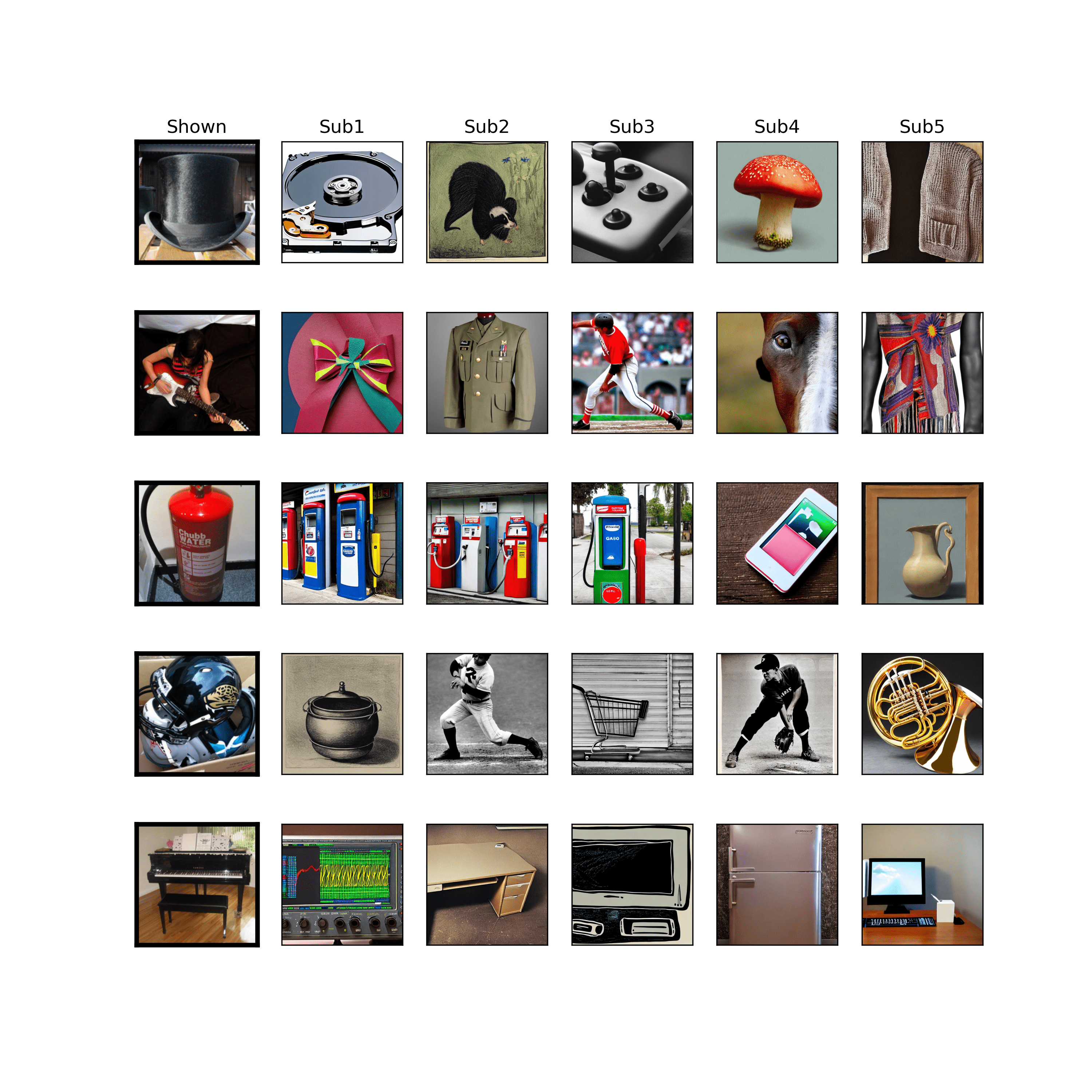}

 \caption{Examples of our semantic reconstructions over the test set. Left columns: original image stimulus shown to the subjects under fMRI. Other columns: semantic reconstructions for each subject in the GOD dataset.}
    \label{fig:examples-app-10}
\end{figure}

\begin{figure}[h!]
\includegraphics[width=1\linewidth]{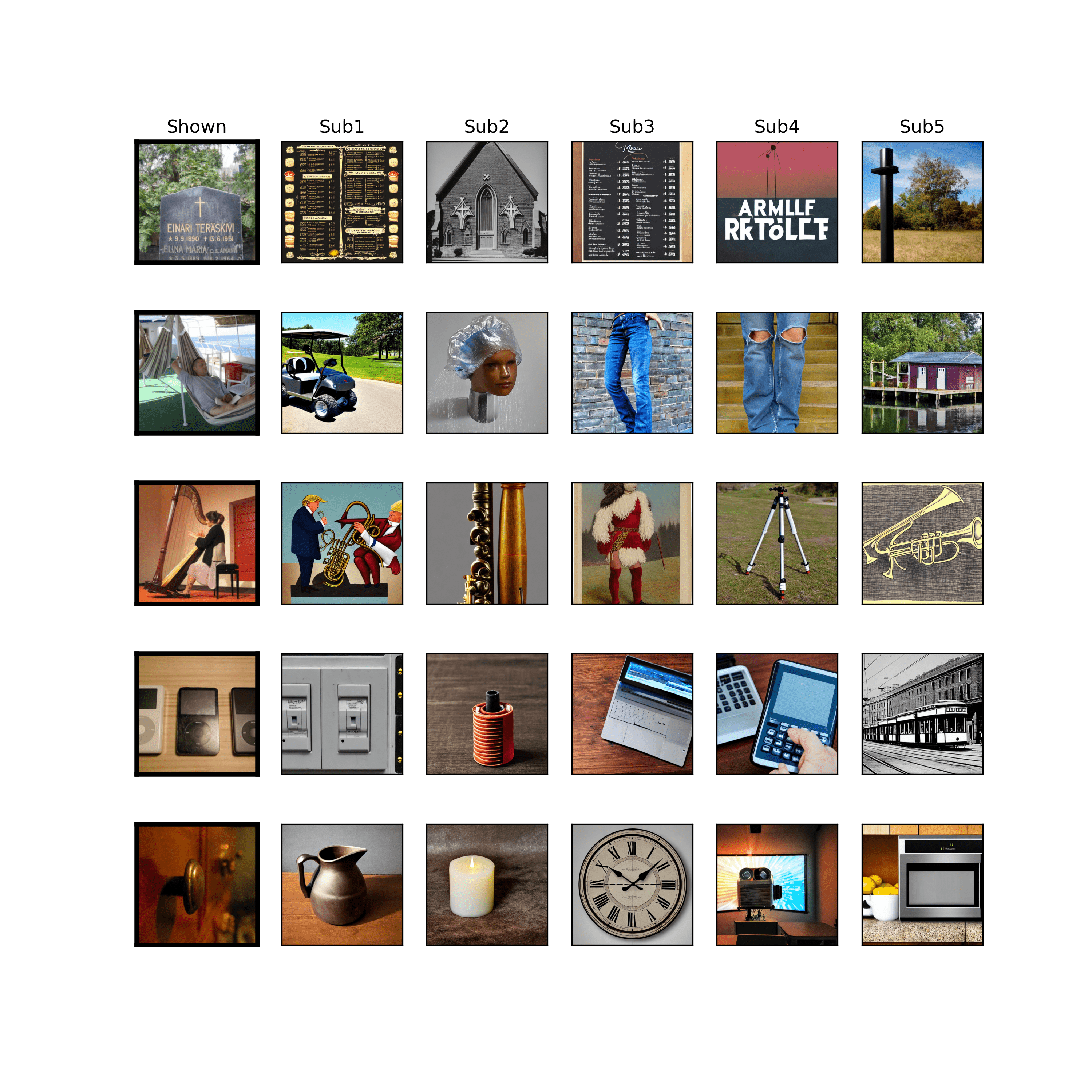}

 \caption{Examples of our semantic reconstructions over the test set. Left columns: original image stimulus shown to the subjects under fMRI. Other columns: semantic reconstructions for each subject in the GOD dataset.}
    \label{fig:examples-app-10}
\end{figure}

\begin{figure}[h!]
\includegraphics[width=1\linewidth]{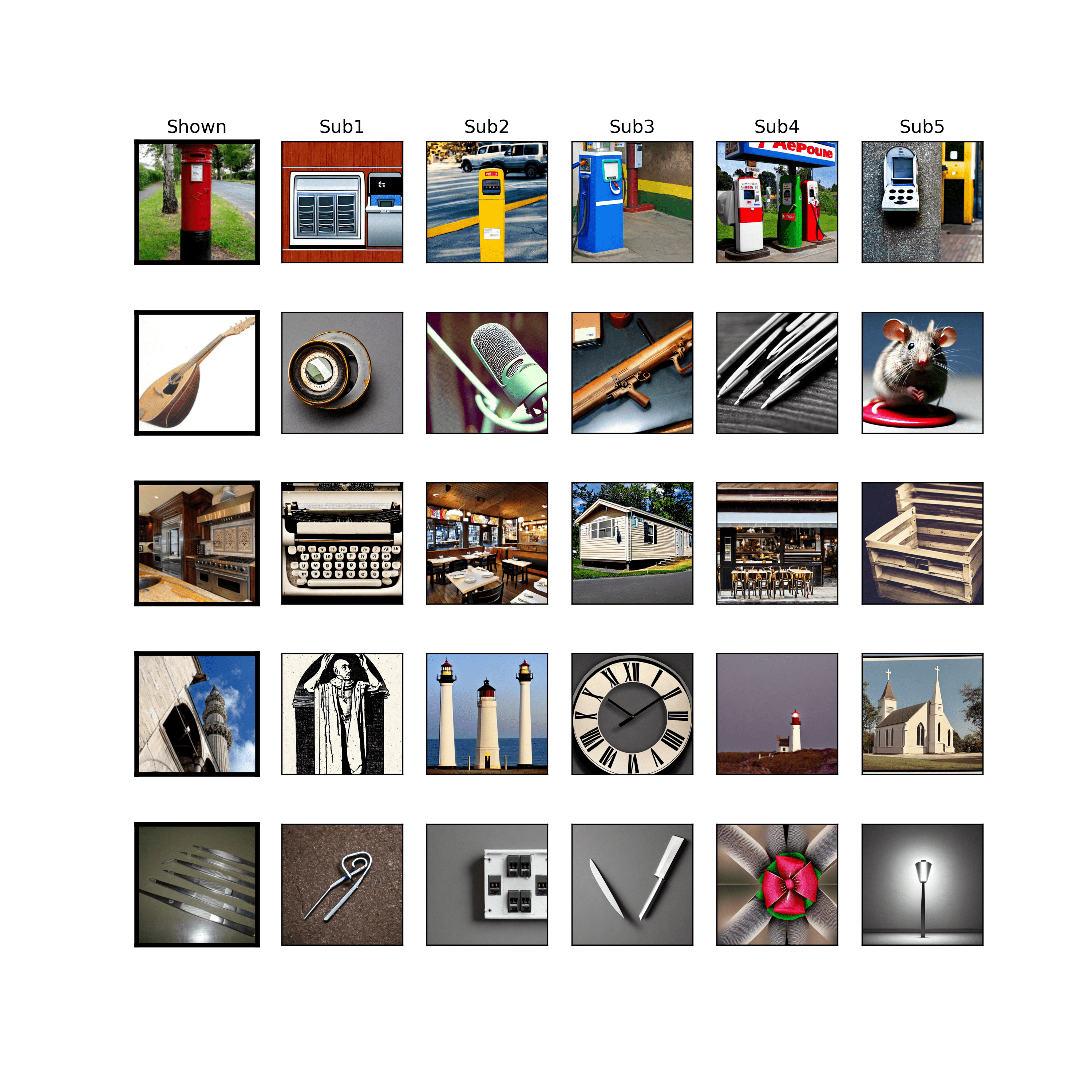}

 \caption{Examples of our semantic reconstructions over the test set. Left columns: original image stimulus shown to the subjects under fMRI. Other columns: semantic reconstructions for each subject in the GOD dataset.}
    \label{fig:examples-app-10}
\end{figure}

\begin{figure}[h!]
\includegraphics[width=1\linewidth]{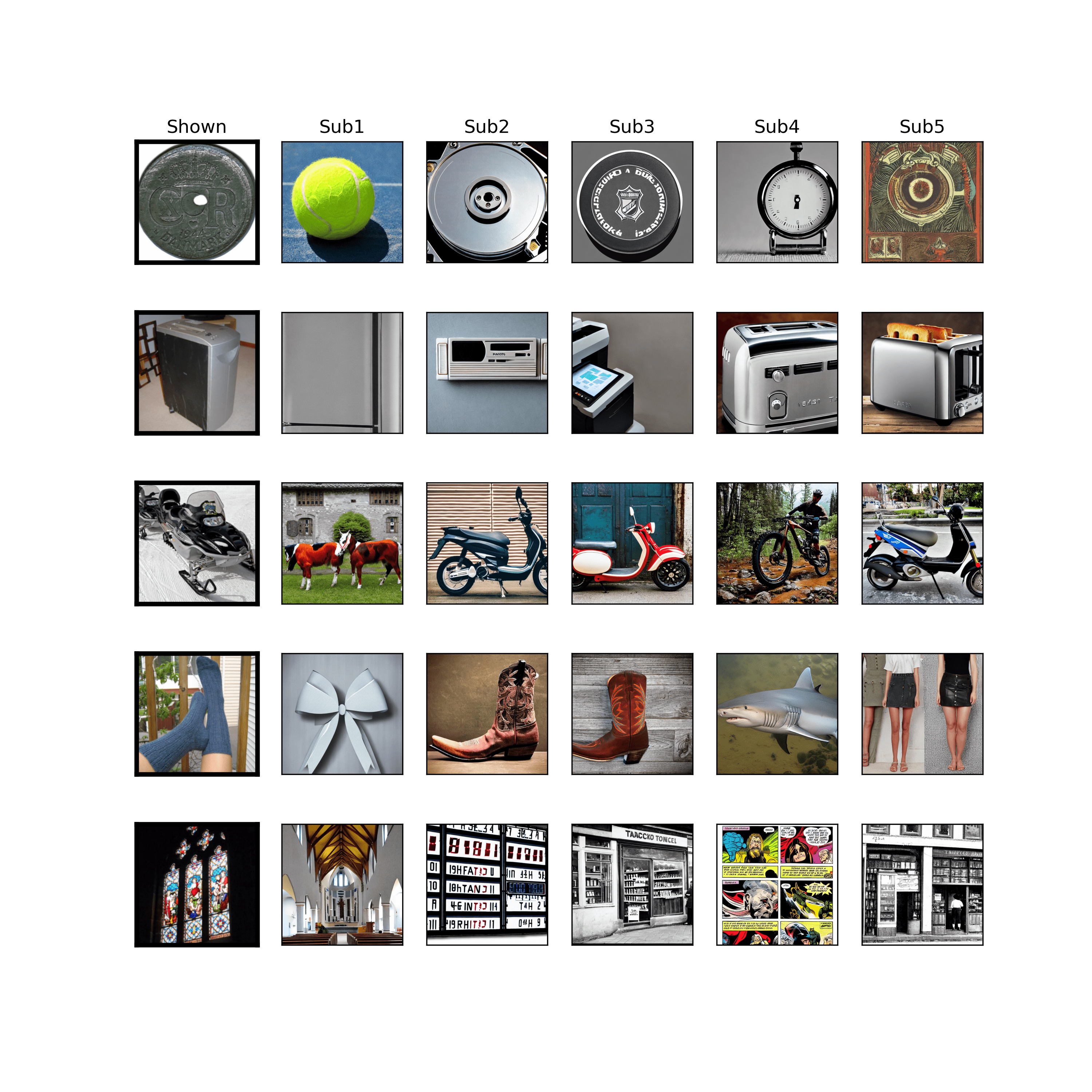}

 \caption{Examples of our semantic reconstructions over the test set. Left columns: original image stimulus shown to the subjects under fMRI. Other columns: semantic reconstructions for each subject in the GOD dataset.}
    \label{fig:examples-app-10}
\end{figure}

\begin{figure}[h!]
\includegraphics[width=1\linewidth]{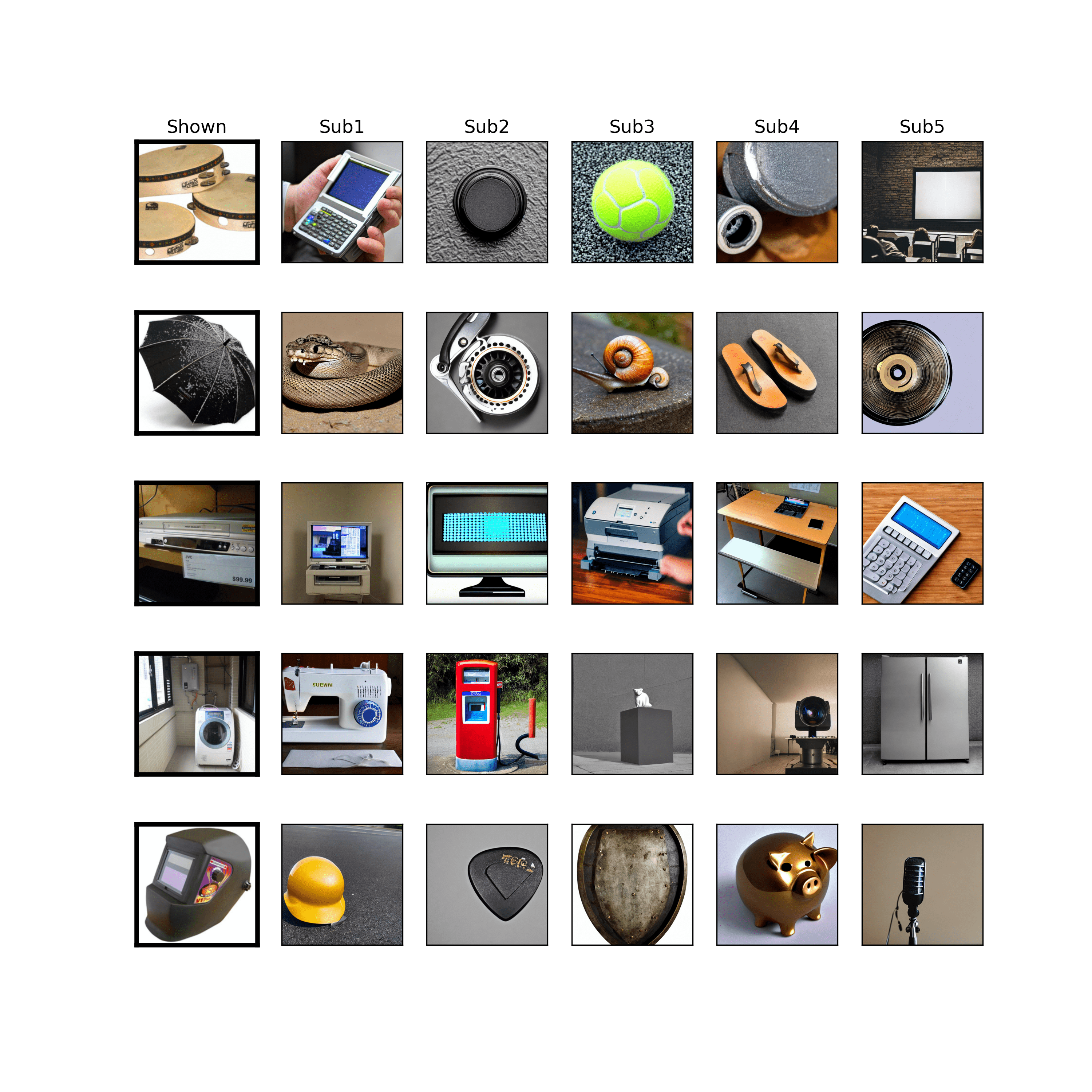}

 \caption{Examples of our semantic reconstructions over the test set. Left columns: original image stimulus shown to the subjects under fMRI. Other columns: semantic reconstructions for each subject in the GOD dataset.}
    \label{fig:examples-app-10}
\end{figure}

\end{document}